\documentclass{article}
\usepackage{spconf,amsmath,graphicx}

\title{A NIR-to-VIS face recognition \\via part adaptive and relation attention module}

\name{Rushuang Xu, MyeongAh Cho, and Sangyoun Lee$^{*}$} 
\address{School of Electrical and Electronic Engineering, Yonsei University, Repulic of Korea\\
\{rushuangxu, maycho0305, syleee\}@yonsei.ac.kr}

\usepackage{multirow, boldline}
\usepackage{amsfonts}
\usepackage{booktabs}
\usepackage{rotating}
\usepackage{tabularx}
\usepackage{xcolor}
\usepackage[caption=false,font=footnotesize]{subfig}

\begin{document}

\maketitle

\begin{abstract}
In the face recognition application scenario, we need to process facial images captured in various conditions, such as at night by near-infrared (NIR) surveillance cameras. The illumination difference between NIR and visible-light (VIS) causes a domain gap between facial images, and the variations in pose and emotion also make facial matching more difficult. Heterogeneous face recognition (HFR) has difficulties in domain discrepancy, and many studies have focused on extracting domain-invariant features, such as facial part relational information. However, when pose variation occurs, the facial component position changes, and a different part relation is extracted. In this paper, we propose a part relation attention module that crops facial parts obtained through a semantic mask and performs relational modeling using each of these representative features. Furthermore, we suggest component adaptive triplet loss function using adaptive weights for each part to reduce the intra-class identity regardless of the domain as well as pose. Finally, our method exhibits a performance improvement in the CASIA NIR-VIS 2.0 \cite{li2013casia} and achieves superior result in the BUAA-VisNir \cite{huang2012buaa} with large pose and emotion variations.
\end{abstract}


\section{Introduction}
\label{intro}
\begin{figure}[t]
	\centering	
	\includegraphics[width=0.9\columnwidth]{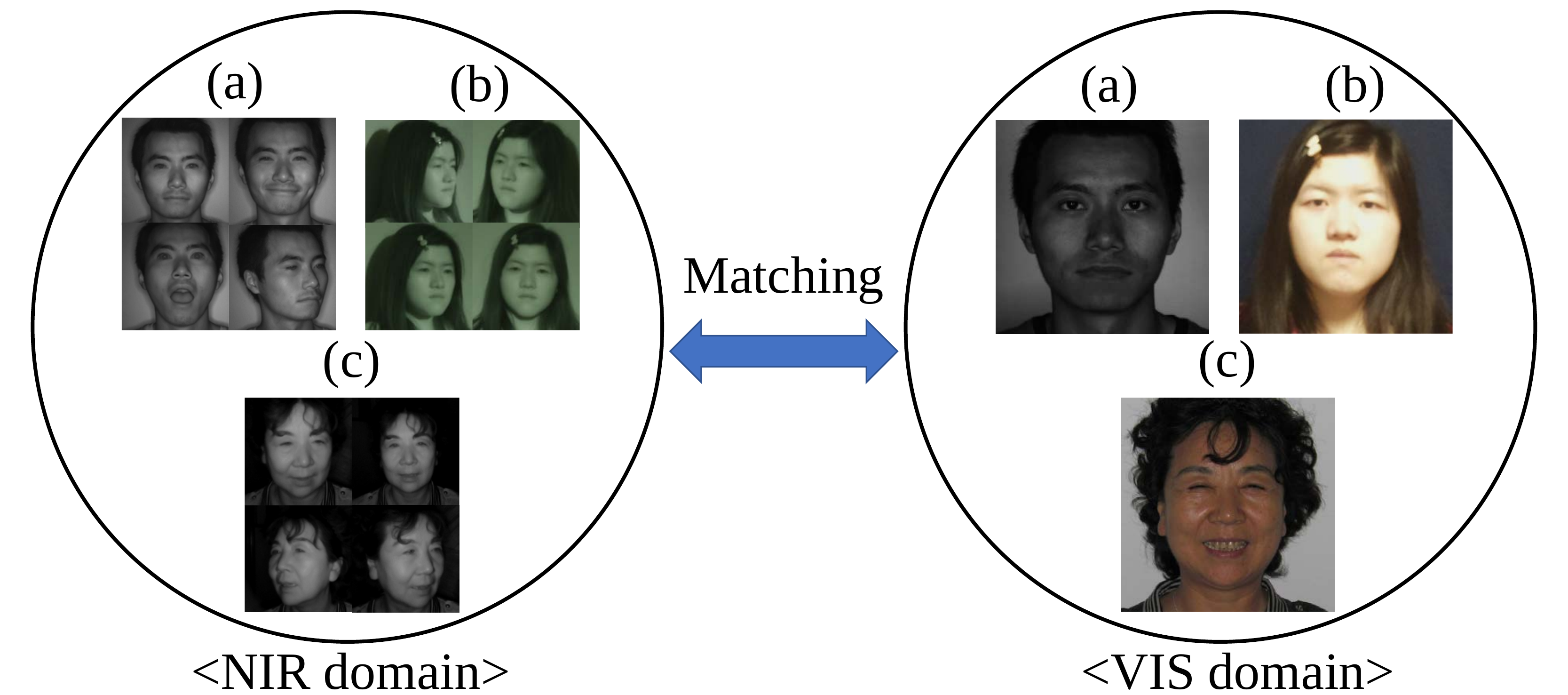}
	\caption{Pose and emotion variation examples of HFR task from (a) BUAA-VisNir \cite{huang2012buaa}, (b) CASIA NIR-VIS 2.0 \cite{li2013casia} and (c) TUFTS \cite{panetta2018comprehensive} face databases.}
	\label{poseex}
\end{figure}

Face recognition is a very common technique in our lives. Not only widespread in daily life, such as in company access control systems, school attendance systems, mobile device unlocking, and so on, but it also plays an important role in assisting public security authorities in handling cases, such as in comparing suspect images captured using surveillance cameras. The near-infrared (NIR) camera used for surveillance systems can capture more useful information at night or in low light conditions than the visible-light (VIS) camera. The task of matching images between the NIR domain and VIS domain is called heterogeneous face recognition (HFR). A domain gap issue is faced by HFR because the NIR images lose considerable spectral information compared to VIS images. Fig. \ref{poseex} reveals that the HFR database faces the challenge of the domain gap and variations in pose and emotion issues. A method to reduce all of these discrepancies is crucial. 

There are two categories of approaches to HFR exist in recent years. One is based on image synthesis methods \cite{wu2019image, song2018adversarial, lezama2017not, bae2020non} that are transferred from one domain (NIR) to another (VIS) using an image synthesis network and matched in a unified domain. However, the quality of the generated images is greatly influenced by the number of training images, which affects recognition performance. The existing HFR database suffers from insufficient data, which is detrimental to the image synthesis method.
Another method is based on learning domain-invariant features. For example, He \textit{et al.}~ \cite{he2018wasserstein} minimized the Wasserstein distance \cite{arjovsky2017wasserstein} between two different domain features, and Liu \textit{et al.}~ \cite{schroff2015facenet} used triplet loss to reduce the domain gap. These methods did not specifically consider issues caused by variations in pose and emotion but simply reduced the overall distance between the two embedding features. 

\begin{figure*}[t]
	\centering	
	\includegraphics[width=0.8\linewidth]{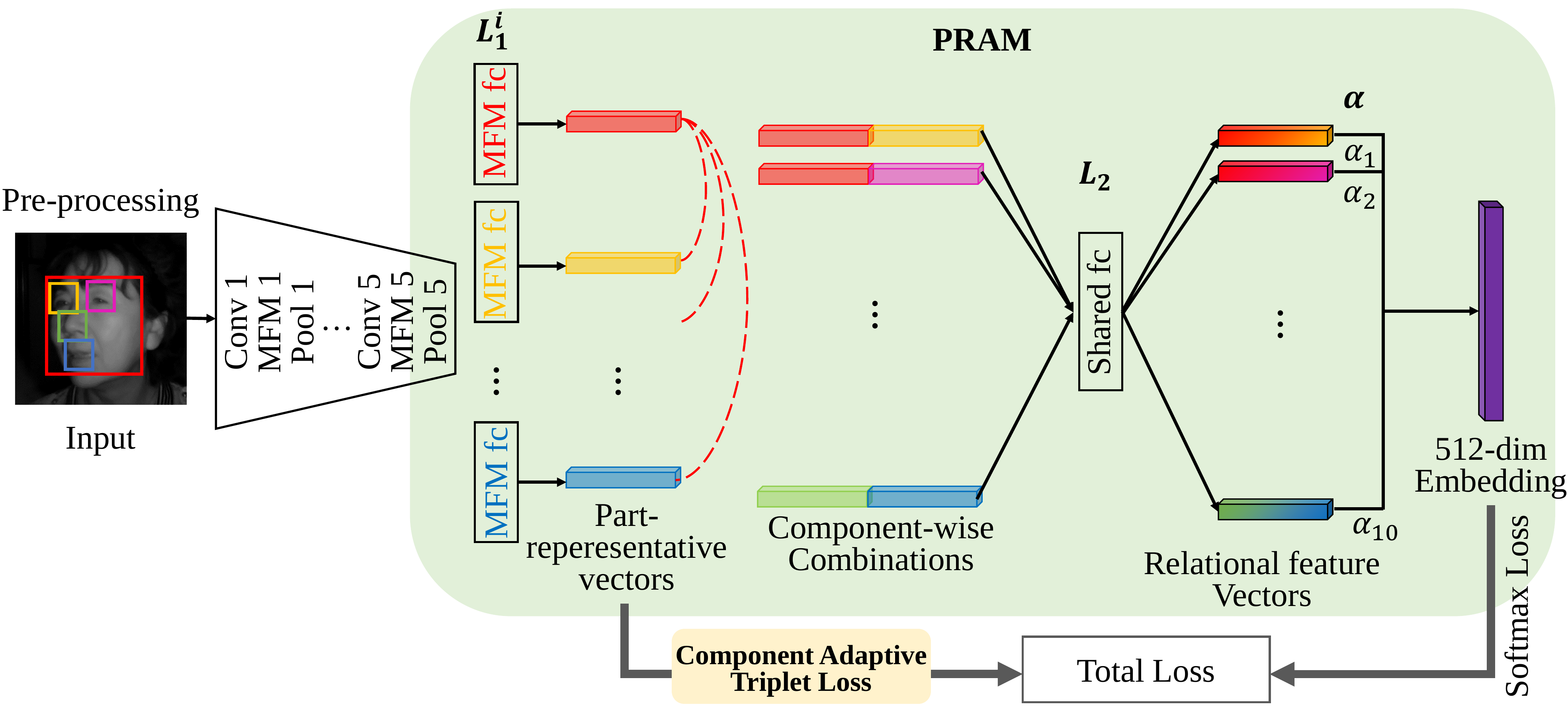}
	\caption{Framework of the proposed model. It extract part representative vectors from the backbone and embedding them into the 512-dim final embedding vector with PRAM. Component adaptive triplet loss and softmax loss are calculated for training}
	\label{framework}
\end{figure*}

We propose a part relation attention module (PRAM) that extracts component relationships based from a facial semantic mask regardless of the domain to learn features robust to the domain and pose and emotion variations. The relationships of the facial components are important for representing domain-invariant identity information.

In addition, we suggest a component adaptive triplet loss function ($L_{CAT}$) that considers pose or emotion variations by assigning adaptive weights according to the visible part of the face. Compared with existing HFR methods, our approach uses relational information among the facial components separated by masking information. Furthermore, the network is trained by considering the part features robust to the pose, emotion, and domain. 

In this paper, our main contributions are as follows:
\begin{itemize}
	\setlength{\itemsep}{0.5em}
	\setlength{\parskip}{0pt}
	\item Uncovering the relationships between facial components (eyes, mouth, and nose) and the entire face, where the proposed PRAM separates these components first then extracts the relations;
	\item Proposing the component adaptive triplet loss function based on a semantic mask to enable the network to learn by selecting information more effectively to solve difficult issues caused by the variation of pose and emotions;
\end{itemize}
\section{Proposed Method}
\label{sec:format}
The overall framework is illustrated in Fig. \ref{framework}. After cropping the facial image into four inputs, the full image and partial image features were extracted and passed through the PRAM layer. For training, the softmax loss is calculated with the embedding feature obtained from the PRAM. Also, the component adaptive triplet loss is calculated with the part-representative features extracted from the backbone. Two terms add up to total loss. While testing, the cosine similarity between the embedding features of each VIS gallery image and the NIR probe image is calculated for recognition. 
\subsection{Part Relation Attention Module (PRAM)}
In HFR, it is important to learn features that irrelevant to the domain such as relational information. Several face recognition studies have been proposed to improve the recognition performance by learning such relations. Chowdhury \textit{et al.}~ \cite{chowdhury2016one} used a bilinear CNN \cite{lin2015bilinear} that multiplied the convolutional layer output feature maps from two-stream CNNs. Cho \textit{et al.}~ \cite{cho2019nir} concatenated the feature vectors of the last feature map pair-wisely to extract the relations between two different facial parts. They both treated each vector of the feature map as representing a certain part of a facial component, such as the lips or nose. However, because the receptive field of the CNN is very large, each vector can almost cover the whole input image. Therefore, these spatially correlated feature vectors have difficulty representing each component separately. In addition, for the images with large pose variations, each feature vector of the same spatial location may not contain the same component of the face. For example, Fig. \ref{poseex} presents the VIS images of frontal view of the face and the NIR images of them are side view of the face. The parts captured in the same location of these images contain different components. Therefore, we design the module to capture the features of each facial part separately to model their relational information.\\
\indent\setlength{\parindent}{1em}As presented in Fig. \ref{framework}, the PRAM comprises four steps. First, a facial image is cropped into four parts: left eye, right eye, nose, and mouth, according to the mask extracted in advance (please refer to the Section \ref{sec3.1}). Unlike landmark detection, since the facial component of the NIR domain is well distinguished by the segmentation network, we use a segmentation mask for parts location. The four partial images and overall facial image (in total, five images) are separately input into the backbone lightCNN-9 \cite{wu2018light}. The first few layers are frozen during training, and only the last layer 8 and 9 are fine-tuned, where the five MFM FC layers \cite{wu2018light} ($L_{1}^{i}$ in Fig. \ref{framework}) are tuned independently. In this step, the representative features of each part are extracted precisely. The MFM FC layer is a special maxout operation that uses a competitive relationship to obtain generalizability, which benefits learning across different data distributions.
In the second step, we extracted the orderless pairwise combinations and arranged them in a fixed order to present the relationship between two parts. In the third step, all combinations are input into a shared FC layer ($L_2$) to guarantee that the network learns the same functional relationship between two representative features. From this computation, the relationship between certain regions with a uniform standard are obtained.
In the last step, we propose a learnable weight \textbf{${\alpha}$} to capture the the strength of each relation. A 512-dim final embedding vector is computed by weighted sum with \textbf{${\alpha}$} of these relational features.

\begin{figure}[t]
	\centering	
	\includegraphics[width=\linewidth]{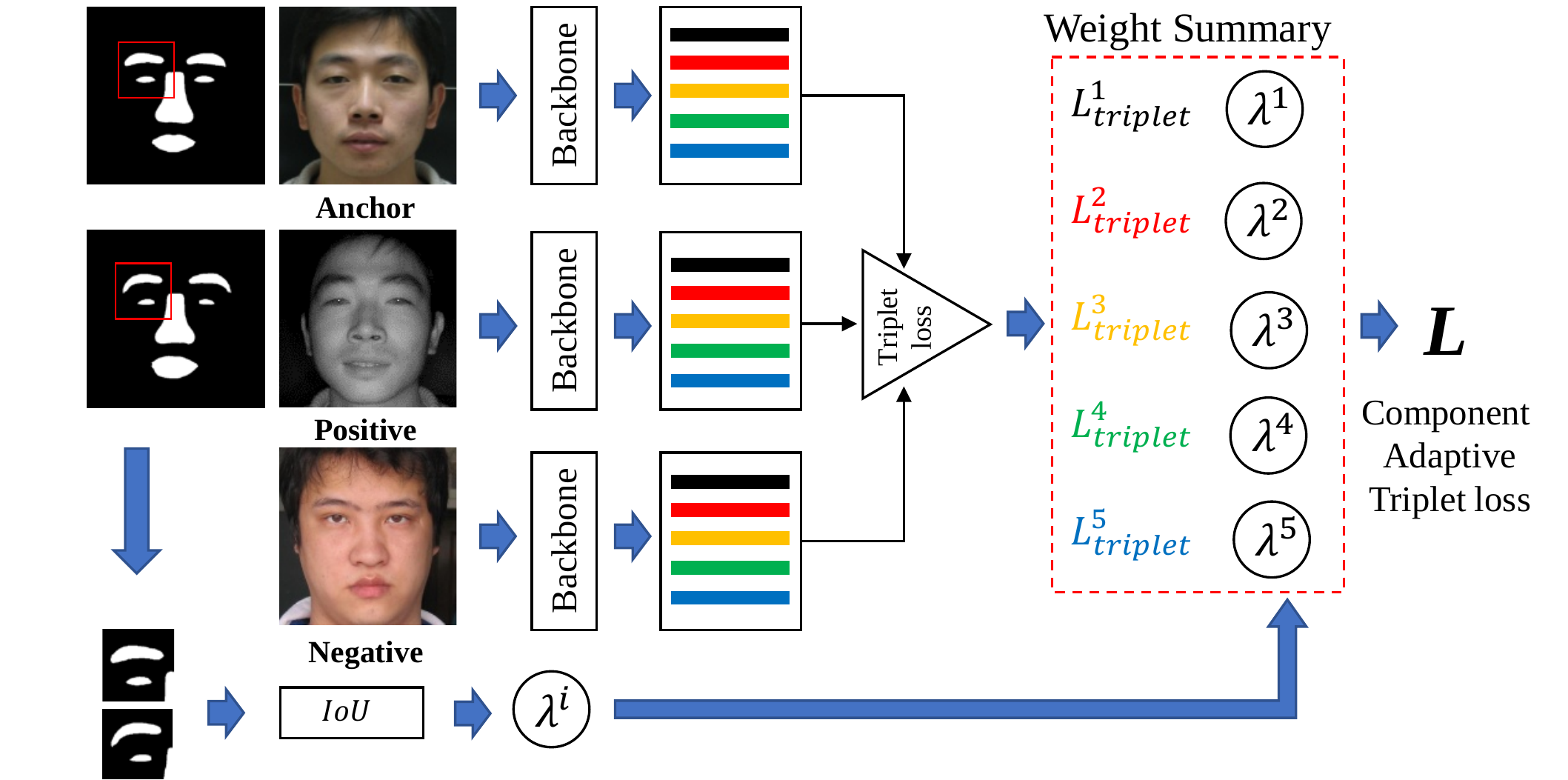}
	\caption{Component adaptive triplet loss structure. The loss value $L_{triplet}^{i}$ is calculated from each component multiplied by the weight $\lambda^{i}$ obtained based on the masking region.}
	\label{loss}
\end{figure}

\subsection{Component Adaptive Triplet Loss Function }
The triplet loss \cite{schroff2015facenet} was proposed to learn more optimized embedding features in latent space by closing the distance between the features of the anchor and positive examples and distancing the features of the anchor and negative examples. The positive example is an image with the same ID as the anchor example, whereas the negative example is different from the anchor. To narrow the intra-class distance between the different domains of the same person strictly, we sample anchor and positive into different domains and negative into the same domain. In addition, an adaptive weight is assigned to the loss of each part-representative vector obtained from the PRAM, considering the different deviations for each component feature caused by the variants of the pose and emotion.

As presented in Fig. \ref{loss}, we use five loss terms for the original image and its four components, separately. In addition, examples exist in the database where some facial features are obscured due to a large pose or emotion variation. To avoid the negative effect of such challenging examples resulting bias in the network training, we propose a loss function that assigns weights adaptively to the five terms based on the matching region between the masks for two examples.
\begin{gather}
\lambda ^{i}=\frac{\sum {M_{a}^{i}\cap M_{p}^{i}}}{\sum {M_{a}^{i}\cup M_{p}^{i}}}
,i=[1,5]
\label{weight}
\end{gather}
The extracted masks of the anchor and positive examples are denoted as $M_{a}^{i}$ and $M_{p}^{i}$. They are both binary images where the background part is set as 0, and the object part is set as 1. In addition, $\lambda^{i}$ is calculated using the intersection over union (IoU), where the above terms are the area of the overlap and the below terms are the area of the union. 
\begin{gather}
S_{p}^{i}=CS(x_{i}^{a},x_{i}^{p})\nonumber\\
S_{n}^{i}=CS(x_{i}^{a},x_{i}^{n})\nonumber\\ 
L_{C}^{i}=[\frac{S_{n}^{i}+1}{S_{p}^{i}+1}-m]_{+}, \hspace{3mm} L_{CAT}^{i}=\sum_{i=1}^{5}\lambda^{i}*L_{C}^{i}
\label{triplet}
\end{gather}
\begin{gather}
L_{total}=s*L_{softmax}+L_{CAT}^{i}
\label{total}
\end{gather}

As for the loss value $L_{C}^{i}$ in Eq. \ref{triplet}, we use the triplet loss with a conditional margin proposed by \cite{cho2020relational} where $CS$ represents the cosine similarity, and the conditional margin $m$ is set to 0.55. The $x_{i}$ indicates feature vector of each component, and $a, p, n$ refer to anchor, positive and negative example, respectively. The final loss function (Eq. \ref{total}) consists of the softmax classification loss with the scaling factor $s=24$ (following \cite{ranjan2017l2}) and the component adaptive triplet loss.

\begin{figure}[t]
	\centering
	\subfloat[VIS]{\includegraphics[width=0.4\columnwidth]{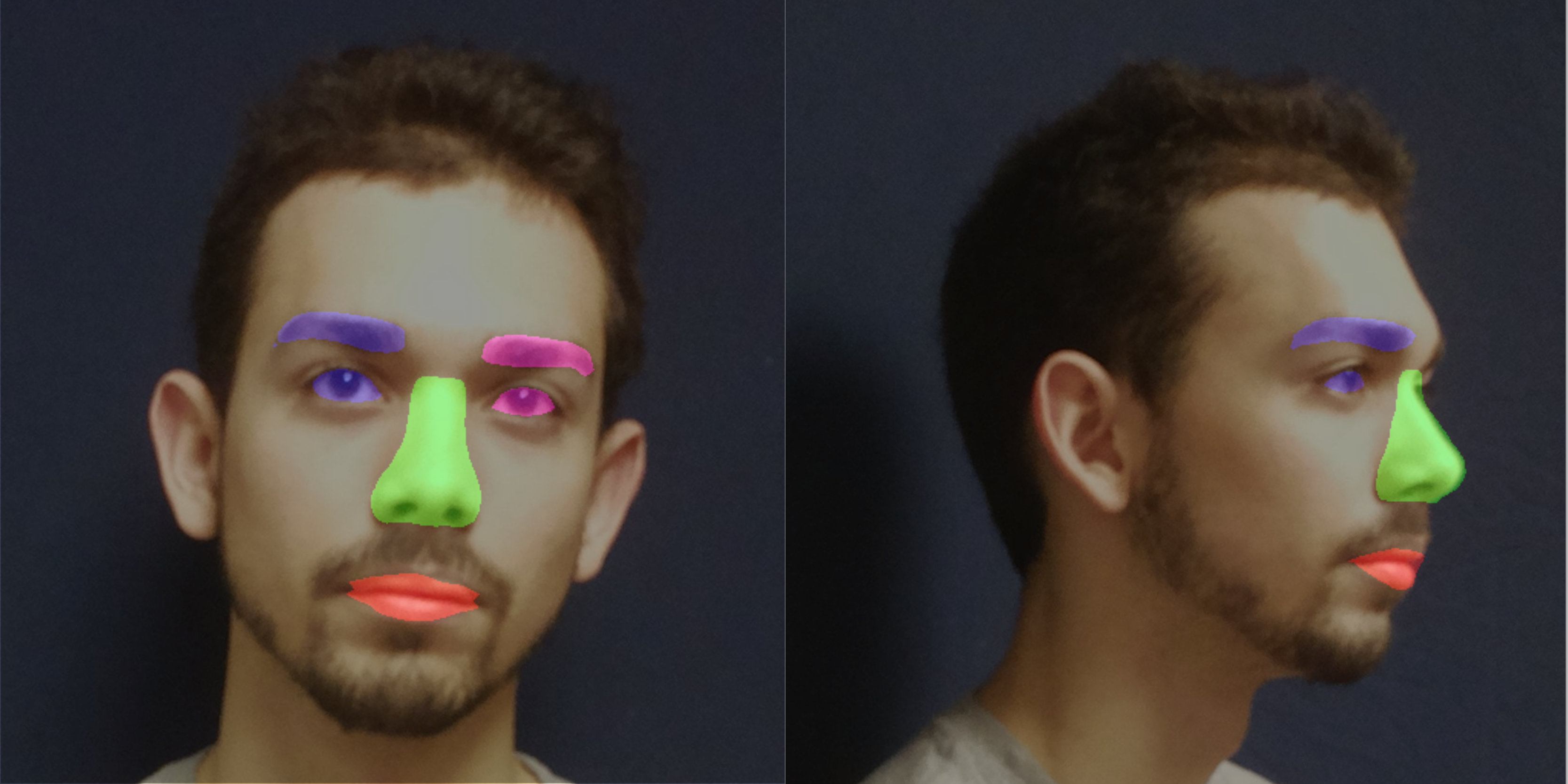}%
		}
		\hfil
	\subfloat[NIR]{\includegraphics[width=0.4\columnwidth]{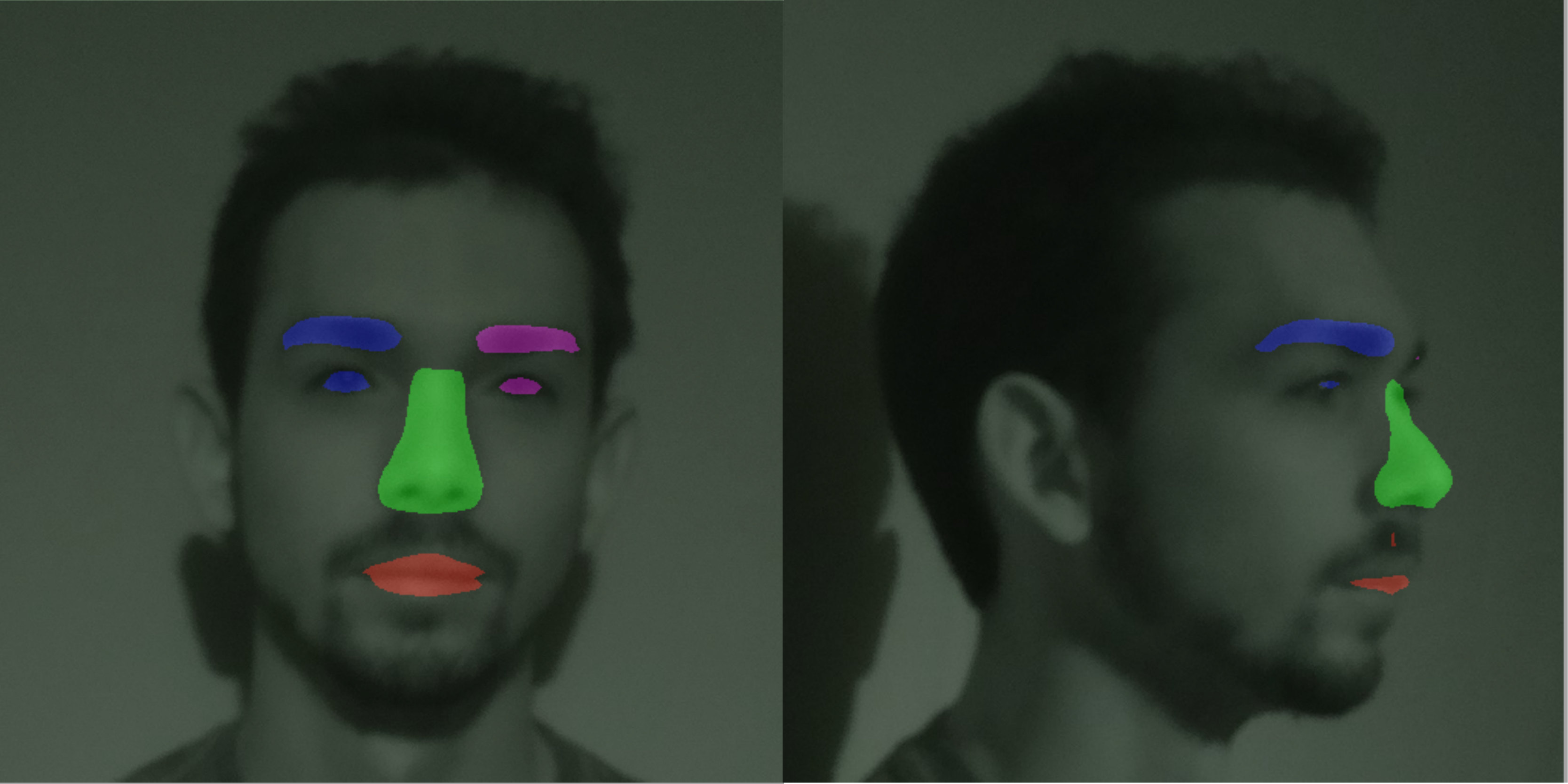}%
		}
	\caption{Results of the semantic segmentation network BiSenet \cite{yu2018bisenet} implemented on the TUFTS face database, used for extracting the mask in our proposed model.}
	\label{TUFTS}
\end{figure}

\begin{table*}[t]
	\centering
	\resizebox{0.85\linewidth}{!}{
	\begin{tabular}{ccccc|cc}
		\hline
		\multicolumn{2}{c|}{\multirow{2}[1]{*}{Models}} &
		\multicolumn{3}{c|}{\textbf{CASIA NIR-VIS 2.0 \cite{li2013casia}}} & \multicolumn{2}{c}{\textbf{BUAA-VisNir \cite{huang2012buaa}}} \\
		\multicolumn{2}{c|}{}    & Rank-1 Acc.(\%)  & VR@FAR=1\%(\%)  & VR@FAR=0.1\%(\%)  & Rank-1 Acc.(\%)    & VR@FAR=1\%(\%)\\ 
		\hline
		\hline
		\multirow{4}[2]{*}{\begin{sideways}LightCNN-9\end{sideways}} & \multicolumn{1}{c|}{Fine-tuned} & 96.06  & 95.19       & 94.06        & 95.77         & 95.88 \\
		\cline{2-7}
		& \multicolumn{1}{c|}{+PRAM} & 97.55  & 95.87       & 95.06        & 96.55         & 95.44 \\
		& \multicolumn{1}{c|}{+ $L_{C}$}    & 98.21  & 97.07       & 96.33        & 98.88         & 97.00 \\
		& \multicolumn{1}{c|}{+ $L_{CAT}$}  & \textbf{98.53}  & \textbf{98.0}       & \textbf{97.49}        & \textbf{99.44}         & \textbf{98.44} \\[0.1cm]
		\hline
	\end{tabular}}
	\caption{Proposed model results on the CASIA NIR-VIS 2.0 and BUAA-VisNir databases}
	\label{t1}
\end{table*}


\section{Experiments}
\label{sec:experiments}
We use the CASIA NIR-VIS 2.0 \cite{li2013casia}, BUAA-VisNir \cite{huang2012buaa}, and TUFTS \cite{panetta2018comprehensive} HFR databases for the experiment. The CASIA NIR-VIS 2.0 database consists of 725 identities, and 8749 images in the training set of 357 identities. The testing set contains 358 identities, and the gallery set has one VIS image for each identity. The probe set has 6208 NIR images. The BUAA-VisNir consists of 150 identities, 900 images from 50 identities for the training set, and 900 images from 100 identities for the testing probe set. For evaluation, the gallery set consists of one VIS image per person. For TUFTS database, which has the largest pose variation, only visualization experiment is conducted since there is no protocol and only composed with 100 ids.

\begin{figure}[t]
	\centering
	\subfloat[BUAA-VisNir]{\includegraphics[width=0.3\columnwidth]{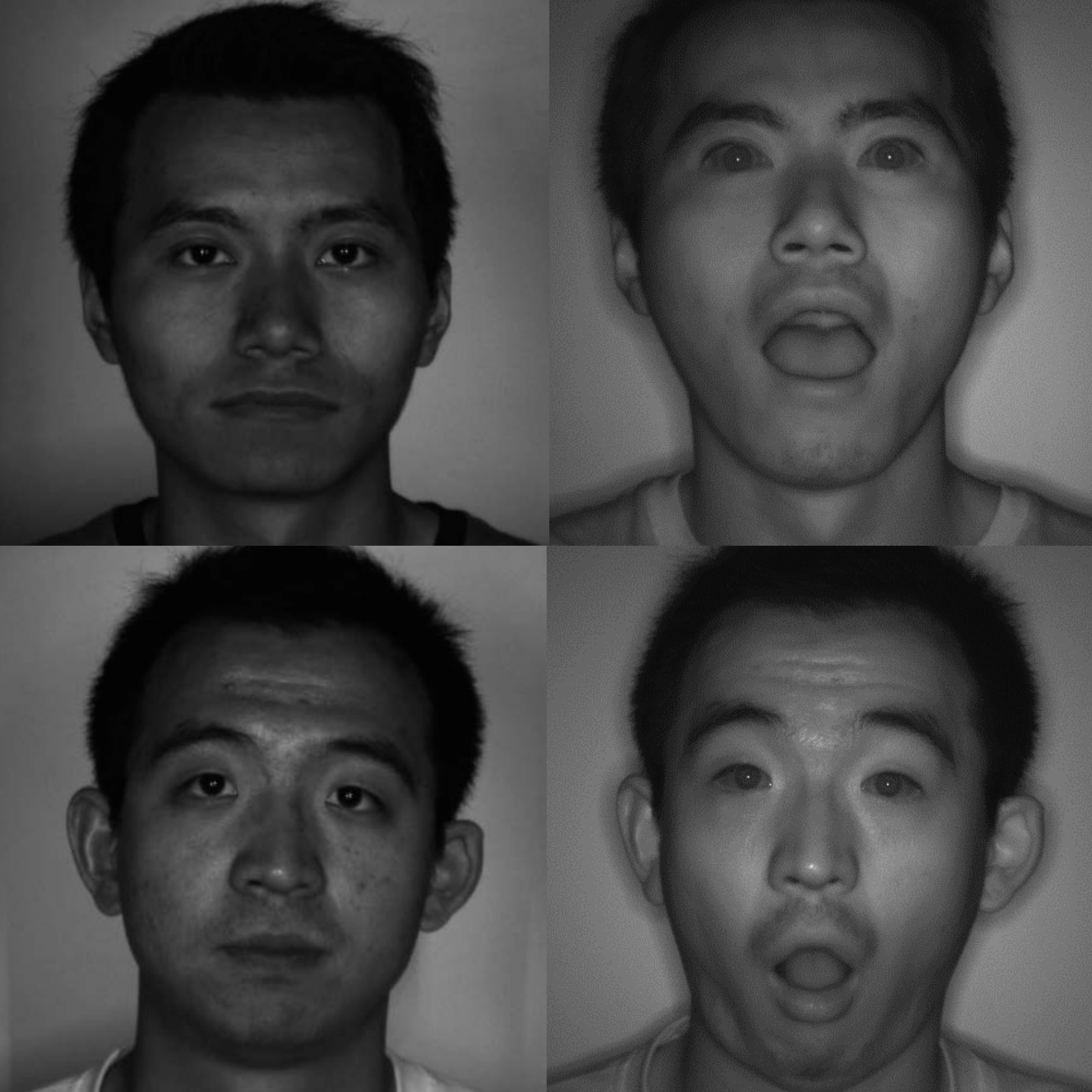}%
		}
		\hfil
	\subfloat[CASIA 2.0]{\includegraphics[width=0.3\columnwidth]{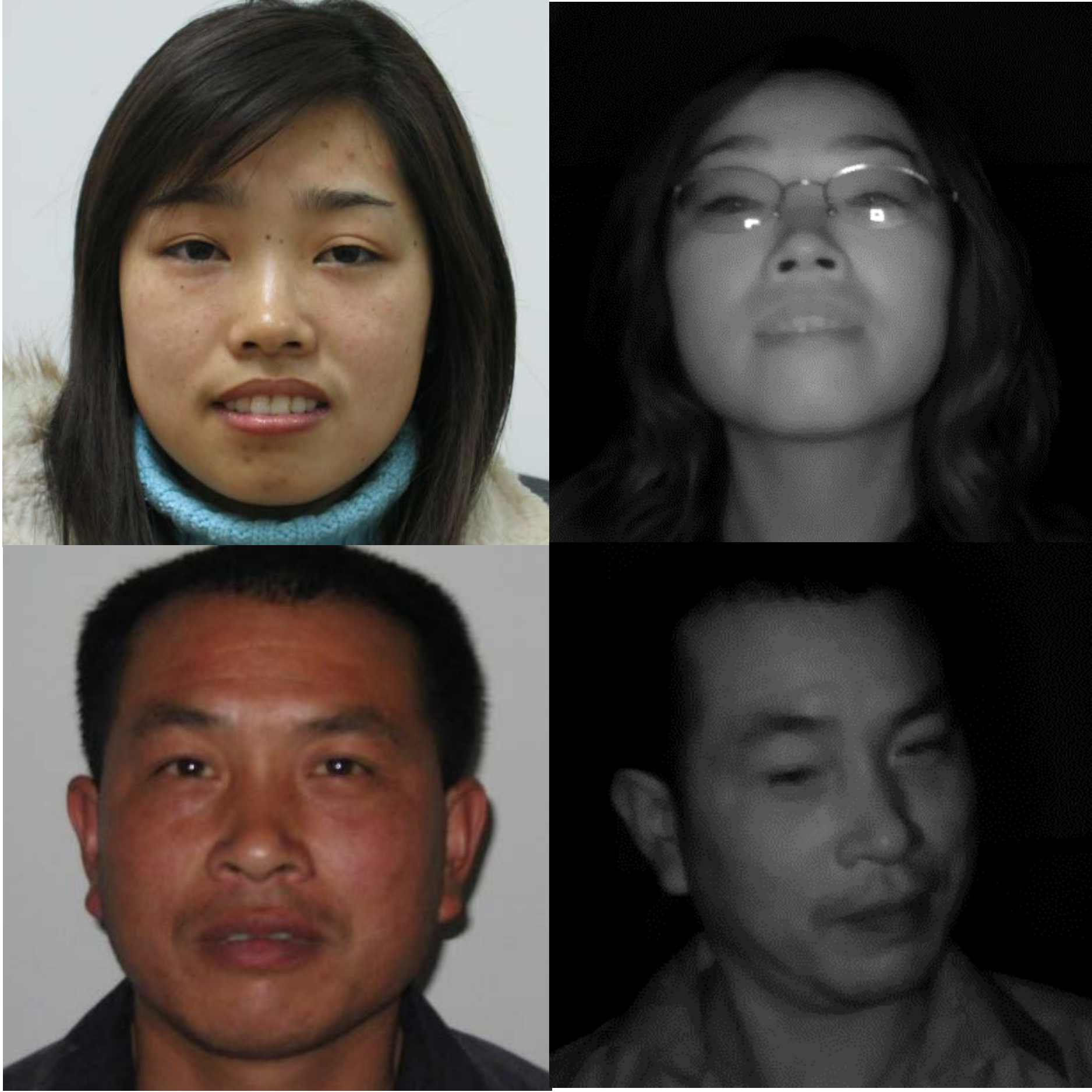}%
		}
		\hfil
	\subfloat[TUFTS]{\includegraphics[width=0.3\columnwidth]{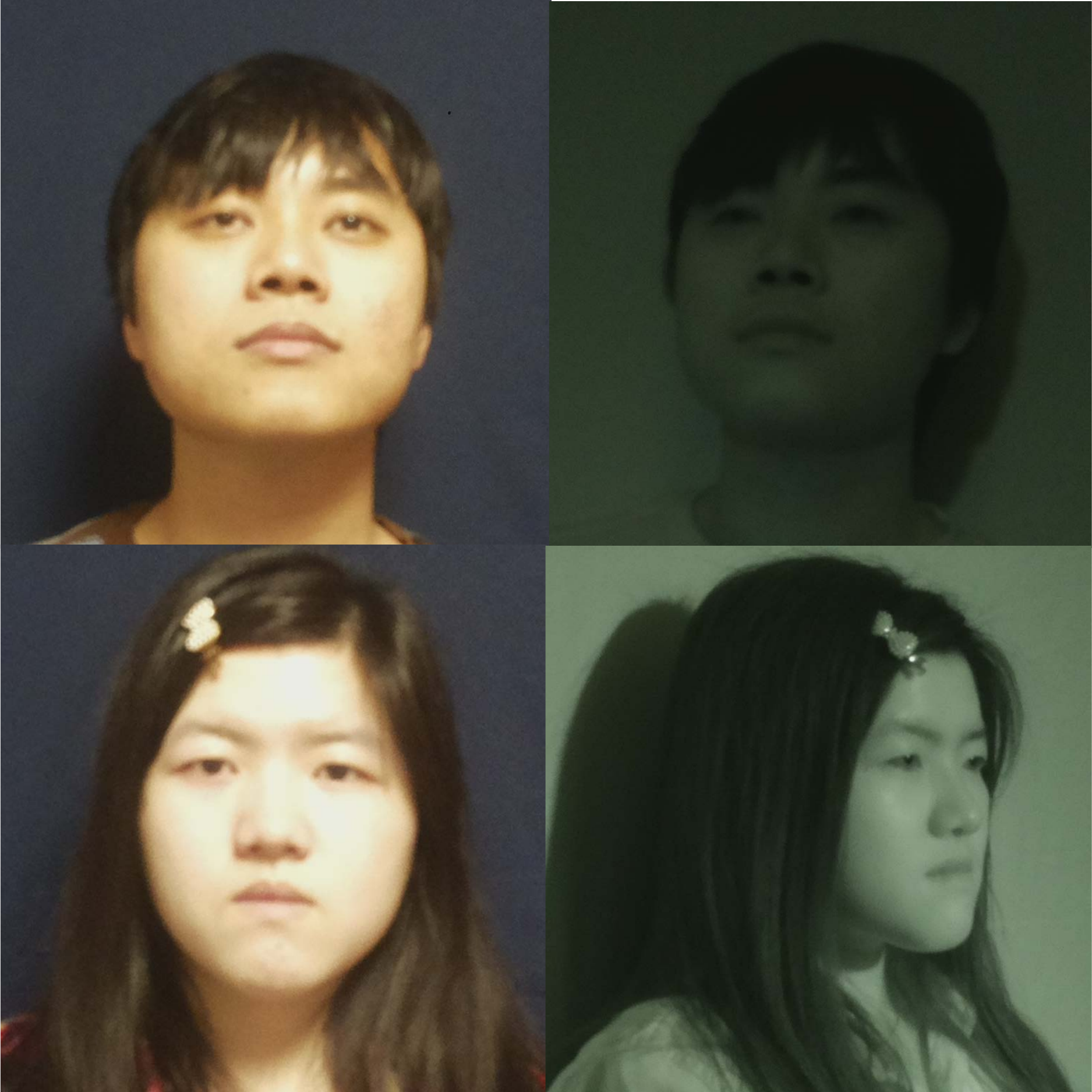}%
		}
	\caption{Successfully recognized samples after employing our proposed model compared with the baseline. In each subfigure, left side is VIS image and the right side is NIR image of same identity.}
	\label{compare}
\end{figure}

\vspace{-3mm}

\subsection{Implementations}
\label{sec3.1}
The masks of facial components are extracted using the real-time semantic segmentation network BiSenet \cite{yu2018bisenet}, which is pretrained on the CelebAMask-HQ database (Fig. \ref{TUFTS}). We crop all the images to 144x144, and randomly crop to 128x128 during the training process. Then, the center point of the rectangle bounding box that contains the mask regions are cropped for the partial images. We use the LightCNN-9 as the baseline which pretrained on the MS-Celeb-1M database \cite{guo2016ms}. We use the stochastic gradient descent optimizer with a learning rate of $10^{-3}$ and a weight decay of $5\times 10^{-4}$. The batch size was set to 16.

\vspace{-3mm}

\subsection{Ablation Studies and Analysis}
In Table \ref{t1}, the Rank-1 accuracy of the baseline on the CASIA NIR-VIS 2.0 and BUAA-VisNir databases are 96.06\% and 95.77\%, respectively. The performance is significantly improved 1.49\% and 0.78\% with the proposed PRAM on both databases. After training with our proposed component adaptive triplet loss for the experiment, the rank-1 accuracy boost to 98.53\%, and 99.44\%. Both results are better than using conditional triplet loss ($L_C$). 

As displayed in Fig. \ref{compare}, we visualize some successfully recognized samples after employing our proposed model compared with the baseline. With the domain discrepancy, pose and emotion variation, the baseline fails to recognize identities in Fig. \ref{compare}. These samples reveal that our model effectively recognizes images with a large variation in emotion and pose.
\subsection{Comparison with Deep Learning Methods}
We compared the our method with other deep learning methods, including TRIVET \cite{liu2016transferring}, IDR \cite{he2017learning}, ADFL \cite{song2018adversarial}, CDL \cite{wu2018coupled}, WCNN \cite{he2018wasserstein}, RM \cite{cho2019nir}, and RGM \cite{cho2020relational}. In Table \ref{t2}, our PRAM performed better than the RM, which pairwise concatenated the feature vector with the addition of conditional triplet loss ($L_C$). In Table \ref{t3}, our approach exhibits the best performance on the BUAA-VisNir database with a large variance in emotion and pose. 
Compared to the WCNN and ADFL, our method is slightly lower in the CASIA NIR-VIS 2.0 database but still demonstrate competitive performance, and higher performance with 2.04\% and 4.24\% in Buaa database. 

\begin{table}[t]
	\centering
	\resizebox{0.75\columnwidth}{!}{
	\begin{tabular}{ccc}
		\hline
		\multicolumn{1}{c}{\multirow{2}[1]{*}{Models}} &
		\multicolumn{2}{c}{\textbf{CASIA NIR-VIS 2.0 \cite{li2013casia}}} \\ 
		\multicolumn{1}{c}{}    & Rank-1 Acc.(\%)  & VR@FAR=0.1\%(\%) \\ 
		\midrule
		\midrule
        TRIVET \cite{liu2016transferring} & 95.7 & 78 \\
        IDR \cite{he2017learning} & 97.33  & 95.73 \\ %
        ADFL \cite{song2018adversarial} & 98.15 & 97.21 \\ %
        CDL \cite{wu2018coupled} & 98.62  & 98.32 \\ %
        WCNN \cite{he2018wasserstein} & 98.7  & 98.4 \\ %
        RM \cite{cho2019nir} & 94.73  & 94.31 \\
        RGM \cite{cho2020relational} & 97.2  & 95.79 \\
        \textbf{Ours} & \textbf{98.53} & \textbf{97.49} \\
		\hline
	\end{tabular}}
	\caption{Comparison with other methods on the CASIA NIR-VIS 2.0 database.\\}
	\label{t2}
\end{table}

\vspace{-3mm}

\begin{table}[t]
	\centering
	\resizebox{0.75\columnwidth}{!}{
	\begin{tabular}{ccc}
		\hline
		\multicolumn{1}{c}{\multirow{2}[1]{*}{Models}} &
		\multicolumn{2}{c}{\textbf{BUAA-VisNir \cite{huang2012buaa}}} \\
		\multicolumn{1}{c}{}  & Rank-1 Acc.(\%)  & VR@FAR=1\%(\%) \\ 
		\midrule
		\midrule
        TRIVET \cite{liu2016transferring} & 93.9  & 80.9 \\
        IDR \cite{he2017learning} & 94.3  & 84.7 \\ 
        ADFL \cite{song2018adversarial} & 95.2  & 95.3 \\
        CDL \cite{wu2018coupled} & 96.9  & 95.9 \\
        WCNN \cite{he2018wasserstein} & 97.4  & 96 \\
        RGM \cite{cho2020relational} & 97.56  & 98.1 \\
        \textbf{Ours} & \textbf{99.44} & \textbf{98.44} \\
		\hline
	\end{tabular}}
	\caption{Comparison with other methods on the BUAA-VisNir Database.}
	\label{t3}
\end{table}
\section{Conclusion}
\label{sec:majhead}
In this paper, we propose a model employed a facial semantic segmentation mask for the location and cropping of facial components and learned domain-invariant features between facial parts through a relational attention structure PRAM. Furthermore, a component adaptive triplet loss function helped efficient learning with large discrepancies in facial parts. We obtained satisfactory performance on HFR databases.
\bibliographystyle{IEEEbib}
\bibliography{ref}
\end{document}